%% file: CameraReady.tex
\documentclass[10pt,twocolumn,letterpaper]{article}

\usepackage{cvpr}              

\usepackage{graphicx}
\usepackage{amsmath}
\usepackage{amssymb}
\usepackage{booktabs}
\usepackage{stfloats} 
\usepackage{xspace} 
\usepackage{amsmath}
\usepackage{multirow}

\usepackage[pagebackref,breaklinks,colorlinks]{hyperref}
\usepackage[dvipsnames]{xcolor}

\ExplSyntaxOn
\newcommand\latinabbrev[1]{
	\peek_meaning:NTF . {
		#1\@}%
	{ \peek_catcode:NTF a {
			#1.\@ }%
		{#1.\@}}} 
\ExplSyntaxOff

\def\eg{\latinabbrev{e.g}}

\def\ie{\latinabbrev{i.e}}

\usepackage[capitalize]{cleveref}
\crefname{section}{Sec.}{Secs.}
\Crefname{section}{Section}{Sections}
\Crefname{table}{Table}{Tables}
\crefname{table}{Tab.}{Tabs.}

\def\cvprPaperID{2433} 
\def\confName{CVPR}
\def\confYear{2023}

\newcommand{\RGBA}{RGB$\alpha$\xspace}

\newcommand*{\affaddr}[1]{#1} 
\newcommand*{\affmark}[1][*]{\textsuperscript{#1}}

\begin{document}

\title{Structural Multiplane Image: Bridging Neural View Synthesis and 3D Reconstruction}

\author{Mingfang Zhang\affmark[1,2*],~~Jinglu Wang\affmark[2],~~Xiao Li\affmark[2],~~Yifei Huang\affmark[1],~~Yoichi Sato\affmark[1],~~Yan Lu\affmark[2]\\
\affaddr{\affmark[1]The University of Tokyo,~~\affmark[2]Microsoft Research Asia}\\
{\tt\small \{mfzhang,hyf,ysato\}@iis.u-tokyo.ac.jp,~~\{jinglwa,xili11,yanlu\}@microsoft.com}
}

\maketitle
\input{Tex/0-abstract.tex}
\input{Tex/1-introduction}

\input{Tex/2-related_works}
\input{Tex/3-representation}

\input{Tex/4-method.tex}

\input{Tex/5-experiment.tex}

\vspace{-1em}
\section{Conclusion}

In this paper, we introduce the Structural MPI (S-MPI) representation, consisting of geometrically-faithful \RGBA images to the scene, for both neural view synthesis and 3D reconstruction. To construct the S-MPI, we propose an end-to-end model in which planar and non-planar regions are uniformly handled. For multi-view input, our model provides a direct global alignment scheme  by generating global proxy embeddings at the full extent of the 3D scene for delivering  aligned images for view synthesis.

\vspace{1em}

\paragraph{Limitation and future work.}
Compared to MPI \cite{zhou2018stereo}, although our Structural MPI achieves better reconstruction and view synthesis results, our method takes more time to construct the S-MPI according to the scene geometry and to render an image because of the intersecting planes. Yet, our rendering process still reaches real-time. For data preparation, we need ground-truth plane segmentation and poses which should be produced from depth map \cite{liu2018planenet}. Finally, in an ideal situation, MPI has the capability to use multiple parallel layers to simulate non-Lambertian effects. However, it is hard for our current method to simulate them as we 
only use one proxy to simulate a surface.
A simple extension can be performed to add more paralleled layers based on our posed planes. This also has advantages over the MPI since we have appropriate orientations to fit the light field. We will leave it for future work.

{\small
\bibliographystyle{ieee_fullname}
\bibliography{egbib}
}

\end{document}

%% file: Tex/0-abstract.tex
\begin{abstract}
   The Multiplane Image (MPI), containing a set of fronto-parallel \RGBA layers, is an effective and efficient representation for view synthesis from sparse inputs. Yet, its fixed structure limits the performance, especially for surfaces imaged at oblique angles. We introduce the Structural MPI (S-MPI), where the plane structure approximates 3D scenes concisely. Conveying \RGBA contexts with geometrically-faithful structures, the S-MPI directly bridges view synthesis and 3D reconstruction. It can not only overcome the critical limitations of MPI, \ie, discretization artifacts from sloped surfaces and abuse of redundant layers, and can also acquire planar 3D reconstruction. Despite the intuition and demand of applying S-MPI, great challenges are introduced, \eg, high-fidelity approximation for both \RGBA layers and plane poses, multi-view consistency, non-planar regions modeling, and efficient rendering with intersected planes. Accordingly, we propose a transformer-based network based on a segmentation model \cite{cheng2021mask2former}. It predicts compact and expressive S-MPI layers with their corresponding masks, poses, and \RGBA contexts. Non-planar regions are inclusively handled as a special case in our unified framework. Multi-view consistency is ensured by sharing global proxy embeddings, which encode plane-level features covering the complete 3D scenes with aligned coordinates. Intensive experiments show that our method outperforms both previous state-of-the-art MPI-based view synthesis methods and planar reconstruction methods.

   \let\thefootnote\relax\footnotetext{*This work was done when Mingfang Zhang was an intern at MSRA.}
\end{abstract}

%% file: Tex/1-introduction.tex
\vspace{-1em}

\section{Introduction}

\begin{figure}[t]
  \centering
   \includegraphics[width=1.0\linewidth]{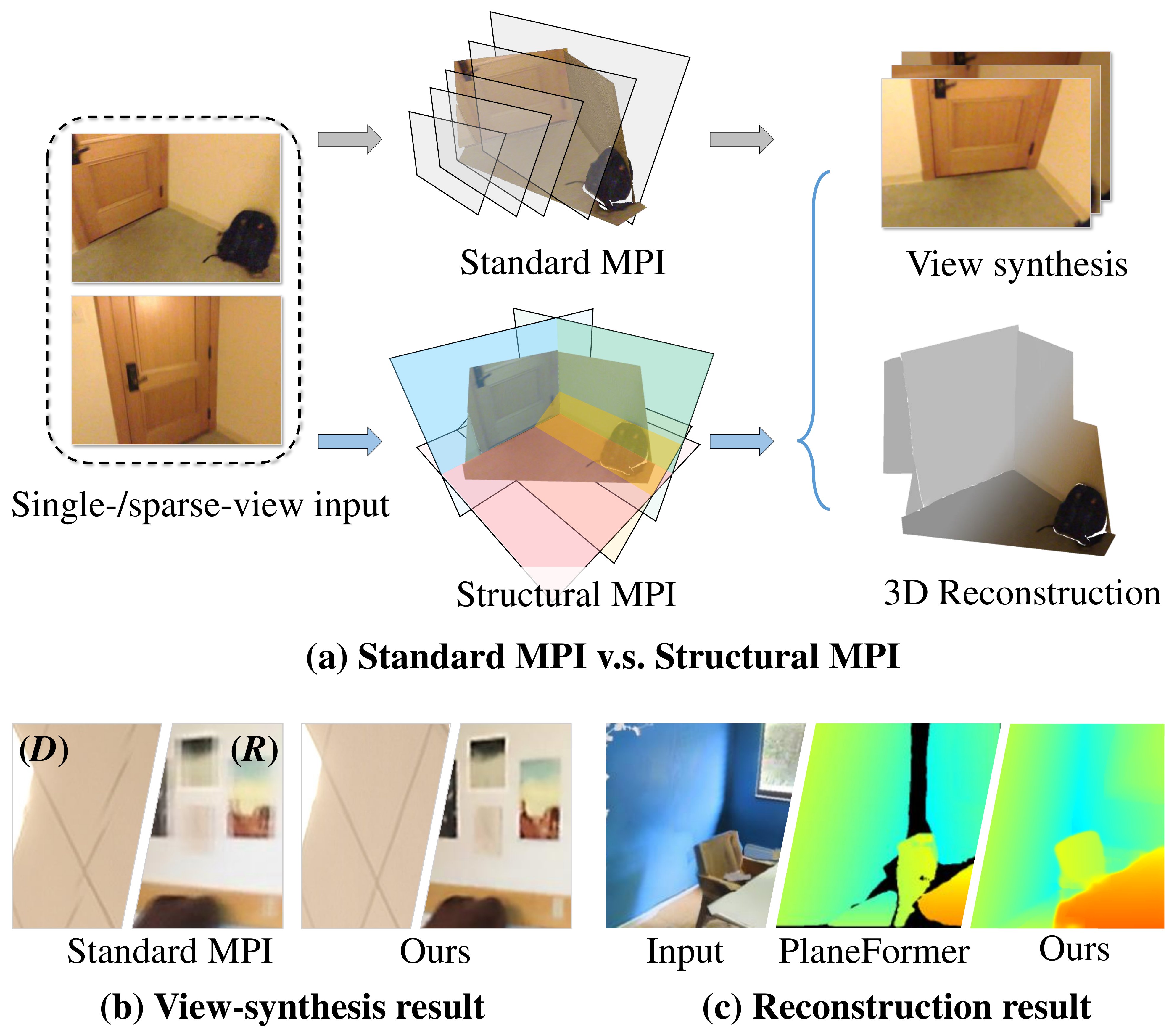}
   \vspace{-2em}
   \caption{
   We propose the Structural Multiplane Image (S-MPI) representation to bridge the tasks of neural view synthesis and 3D reconstruction. It consists of a set of posed \RGBA images with geometries approximating the 3D scene. The scene-adaptive S-MPI overcomes the critical limitations of standard MPI \cite{mine2021}, \eg, \textit{discretization artifacts (D)} and \textit{repeated textures (R)}, and achieves a better depth map compared with the previous planar reconstruction method, PlaneFormer \cite{agarwala2022planes}.
   }
   \label{fig:teasor}
   \vspace{-1.5em}
\end{figure}

Novel view synthesis \cite{zhou2016view, mildenhall2020nerf} aims to generate new images from specifically transformed viewpoints given one or multiple images. It finds wide applications in augmented or mixed reality for immersive user experiences.

The advance of neural networks mostly drives the recent progress. NeRF-based methods \cite{mildenhall2020nerf, martin2021nerf} achieve impressive results but are limited in rendering speed and generalizability. 
The multiplane image (MPI) representation \cite{zhou2018stereo,tucker2020single} shows superior abilities in these two aspects, especially given extremely sparse inputs. Specifically, neural networks are utilized to construct MPI layers, containing a set of fronto-parallel \RGBA planes regularly sampled in a reference view frustum. 
Then, novel views are rendered in real-time through simple homography transformation and integral over the MPI layers.
Unlike NeRF models, MPI models do not need another training for a new scene.

Nevertheless, standard MPI has underlying limitations. 
1) It is sensitive to discretization due to slanted surfaces in scenes. As all layered planes are parallel to the source image plane, slanted surfaces will be distributed to multiple MPI layers causing discretization artifacts in novel views, as shown in \cref{fig:teasor} (b). Increasing the number of layers can improve the representation capability \cite{srinivasan2019pushing} but also increase memory and computation costs.
2) It easily introduces redundancy. 
It tends to distribute duplicated textures into different layers to mimic the lighting field \cite{li2020synthesizing}, which can introduce artifacts with repeated textures as shown in \cref{fig:teasor} (b). 
The essential reason causing the above issues is that the MPI construction is dependent on source views but neglects the explicit 3D geometry of the scenes.
Intuitively, we raise a question: Is it possible to construct MPIs adaptive to the scenes, considering both depths and orientations?

Slanted planes are smartly utilized in 3D reconstruction to save stereo matching costs \cite{gallup2007real} or represent the scene compactly \cite{ikehata2015structured,gallup2010piecewise}, especially for man-made environments.
Recent advanced neural networks reach end-to-end planar reconstruction from images by formulating it as instance segmentation \cite{liu2018planenet,liu2022planemvs}, where planes are directly detected and segmented from the input image. However, non-planar regions are often not well-modeled or neglected \cite{agarwala2022planes,liu2018planenet,tan2021planetr}, resulting in holes or discontinuities in depth maps,
as shown in \cref{fig:teasor} (c). 

In this paper, we aim to bridge the neural view synthesis and the planar 3D reconstruction, that is, to construct MPIs adaptive to 3D scenes with planar reconstruction and achieve high-fidelity view synthesis.
The novel representation we propose is called \textbf{Structural MPI} (S-MPI), which is fully flexible in both orientations and depth offsets to approximate the scene geometry, as shown in \cref{fig:teasor} (a). 
Although our motivation is straightforward, there are great challenges in the construction of an S-MPI.
(1) The network not only needs to predict \RGBA values but also the planar approximation of the scene. 
(2) It is difficult to correspond the plane projections across views since they may cover different regions and present different appearances.
Recent plane reconstruction works \cite{jin2021planar, agarwala2022planes} build matches of planes after they are detected in each view independently, which may increase costs and accumulate errors.
(3) Non-planar regions are challenging to model even with free planes. Previous plane estimation methods \cite{xie2022planarrecon,tan2021planetr,agarwala2022planes} cannot simultaneously handle planar and non-planar regions well.
(4) In the rendering process, as an S-MPI contains planes intersecting with each other, an efficient rendering pipeline needs to be designed so that the rendering advantages of MPI can be inherited.

To address these challenges, we propose to build an S-MPI with an end-to-end transformer-based model for both planar geometry approximation and view synthesis, in which planar and non-planar regions are processed jointly.
We follow the idea \cite{liu2018planenet} of formulating plane detection as instance segmentation and leverage the segmentation network \cite{cheng2021mask2former}. Our S-MPI transformer uniformly takes planar and non-planar regions as two structure classes and predicts their representative attributes, which are for reconstruction (structure class, plane pose and plane mask) and view synthesis (\RGBA image). We term each instance with such attributes as a \textit{proxy}.
Note that non-planar layers are inclusively handled as fronto-parallel planes with adaptive depth offsets and the total number of the predicted proxy instances is adaptive to the scene. 

Our model can manipulate both single-view and multi-view input. It aims to generate a set of proxy embeddings in the full extent of the scene, covering all planar and non-planar regions aligned in a global coordinate frame. For multi-view input, the proxy embeddings progressively evolve to cover larger regions and refine plane poses as the number of views increases.
In this way, the predicted proxy instances are directly aligned, which avoids the sophisticated matching in two-stage methods \cite{agarwala2022planes, jin2021planar}. The global proxy embeddings are effectively learned with the ensembled supervision from all local view projections. 
Our model achieves state-of-the-art performance for single-view view synthesis ($10\%\uparrow$ PSNR) and planar reconstruction ($20\%\uparrow$ recall) in datasets \cite{dai2017scannet,silberman2012indoor} of man-made scenes and also achieve encouraging results for multi-view input compared to NeRF-based methods with high costs.

In summary, our main contributions are as follows:
\begin{itemize}
\vspace{-0.8em}
\item We introduce the Structural MPI representation, consisting of geometrically-faithful \RGBA images to the 3D scene, for both neural view synthesis and 3D reconstruction.
\vspace{-0.8em}
\item{We propose an end-to-end network to construct S-MPI, where planar and non-planar regions are uniformly handled with high-fidelity approximations for both geometries and light filed.  
}
\vspace{-0.8em}
\item {
Our model ensures multi-view consistency of planes by introducing the global proxy embeddings comprehensively encoding the full 3D scene, and they effectively evolve with the ensembled supervision from all views.
}
\end{itemize}

%% file: Tex/2-related_works.tex
\vspace{-0.5em}

\section{Related Works}

\vspace{-0.5em}

\paragraph{View synthesis with explicit representations.} Various methods are proposed for novel view synthesis based on different representations, such as point cloud \cite{wiles2020synsin} and mesh \cite{hu2021worldsheet}.
Layered representations have been the subject of people's interest. Layered Depth Image (LDI) \cite{shade1998layered,shih20203d, tulsiani2018layer, dhamo2019peeking} uses several layers of depth maps and associated color values to represent a scene. Multiplane Image (MPI) \cite{zhou2018stereo} is a popular variant of LDI, in which layer depths are fixed and an alpha channel is introduced. \cite{tucker2020single} proposes a model to construct an MPI from single-view images and \cite{flynn2019deepview, mildenhall2019llff} work with densely sampled multi-view image input. \cite{srinivasan2019pushing} increases the number of layers to enhance MPI's capacity, while \cite{li2020synthesizing, han2022single} claims that adding planes will make the over-parameterized problem worse and the network tends to repeat the content over multiple planes because of depth uncertainty. \cite{han2022single, luvizon2021adaptive} propose to solve the problem by placing planes at adaptive depths. We propose the Structural MPI that overcomes the MPI's drawbacks like discretization artifacts from sloped surfaces and abuse of redundant layers.

\vspace{-1em}

\paragraph{View synthesis with implicit representations.} Implicit representations are popular for view synthesis recently, such as NeRF \cite{martin2021nerf, yu2021pixelnerf, mildenhall2020nerf}, which encodes 3D objects and scenes in the weights of an MLP. A recent method \cite{lin2022neurmips} leverages a collection of planar experts in NeRF, but it relies on pre-acquired point clouds.
Although NeRF-based methods can achieve promising view synthesis results, they are limited in rendering speed and generalizability.
People have tried to develop various solutions \cite{trevithick2021grf, deng2022depth, xu2022sinnerf, muller2022instant, sun2022direct, chen2021mvsnerf, johari2022geonerf} and our method has natural advantages, especially with single-view input and no finetuning session. We can achieve comparable results as NeRF with sparse input views while rendering an image significantly faster than NeRF methods.

\vspace{-1em}

\paragraph{Planar 3D Approximation.} MPI uses a set of fronto-parallel planes to model the scene while planes in a scene have various orientations. Piece-wise planar depth map reconstruction has been a traditional research topic. 
\cite{sinha2009piecewise, gallup2010piecewise} reconstruct 3D points and perform plane-fitting and recent learning-based methods directly detect and reconstruct 3D planes. \cite{liu2018planenet,liu2019planercnn} generates plane segments and 3D plane parameters for the ScanNet dataset \cite{dai2017scannet} and proposes a detection-based framework. \cite{yu2019single, tan2021planetr} learns an embedding for each pixel and groups them to generate plane instances. 
\cite{agarwala2022planes,xie2022planarrecon,jin2021planar} solve the problem with multi-view images and conduct plane matching and fusion to generate corresponding instances. Our method generates plane reconstructions with correspondence directly without a second-stage matching and predicts \RGBA contents on each plane.

%% file: Tex/3-representation.tex
\section{Structural Multiplane Image}
In this section, we introduce the structural multiplane image representation by first elaborating on the geometry formulation based on standard MPI formulation, and then detailing the process of rendering.

\subsection{Geometry Formulation}
\label{sec:mpi-geo-form}
\paragraph{MPI preliminaries.}
The standard MPI consists of a collection of $N$ planes parallel to the image plane,
$\mathcal{P}_c = \{ (C_i, A_i, d_i) \}, i=1, ..., N$, where $C_i \in \mathbb{R}^{H\times W \times 3}$ and $A_i \in \mathbb{R}^{H\times W \times 1}$ denote RGB and alpha transparency maps of size $H \times W$, and $d_i$ denotes the plane offset to the optical center $\mathbf{o}$.

\vspace{-1em}

\paragraph{S-MPI formulation.} Differently, our S-MPI contains a set of \RGBA images on $N_p$ plane layers with their geometries faithful to the scene, \ie, $\mathcal{P}_s = \{ (C_i, A_i, \mathbf{\pi}_i)\}_{i=1}^{N_p}$. The number of planes $N_p$ is adaptive.
The plane geometry $\mathbf{\pi}=(\mathbf{n}, d)$ is represented by a normal vector $\mathbf{n}$ and an offset scalar $d$. Note that each 3D point $\mathbf{x}$ on the plane satisfies:
\vspace{-0.8em}
\begin{equation}
\vspace{-0.8em}
  \label{eq:plane_eq}
    \mathbf{n} \cdot \mathbf{x} = d.
\end{equation}
Then, each proxy of the S-MPI is represented as $(C_i, A_i, \mathbf{n}_i, d_i)$.
\vspace{-1em}
\paragraph{Non-planar region.} 
There are often some non-planar regions not suitable to fit large planes. We consider them not necessary to be forced to fit into small fragmented planes, which can increase fitting errors and rendering costs. Luckily, the fronto-parallel MPI is a special case of S-MPI where all planes are with the same normal as the image plane, \ie, $\mathbf{n}_z=(0, 0, 1)$.
We utilize the inclusive property to simply distribute non-planar regions into nearby fronto-parallel planes as standard MPI. Yet, the offsets $d$ can be adaptive to the depth distribution of the scenes. We then get the S-MPI representation for non-planar regions as $\{(C_i, A_i, \mathbf{n}_z, d_i)\}_{i=1}^{N_n}$. 

Therefore, our S-MPI contains hybrid structures, \ie, geometrically faithful planes approximating scenes' planar regions and depth-adaptive fronto-parallel planes for non-planar regions. Despite the differences in structures, the two types of multiplane images can be unified in our S-MPI formulation. The final S-MPI for the complete scene is extended as $\mathcal{P}_s = \{ (C_i, A_i, \mathbf{n}_i, d_i)\}_{i=1}^{N_p+N_n}$, where $N_n$ elements are with a fixed normal $\mathbf{n}_z$.

\begin{figure}[t]
  \centering
   \includegraphics[width=1.0\linewidth]{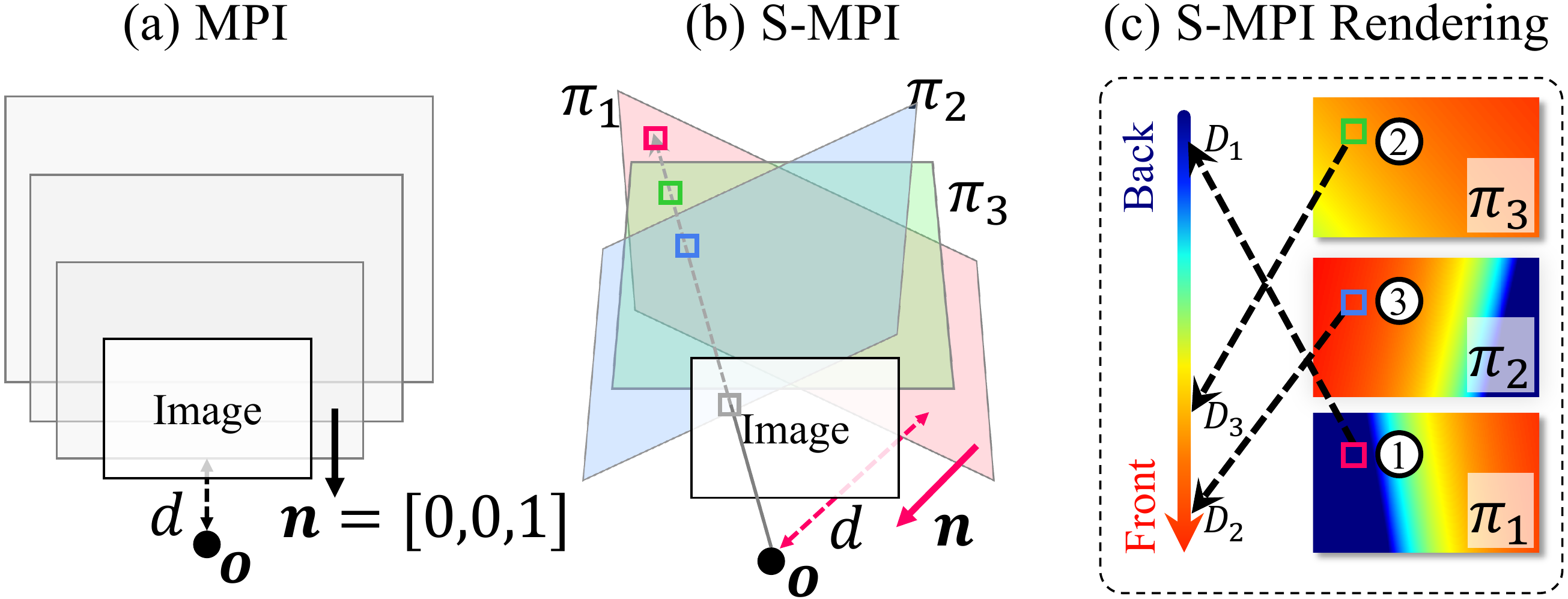}
   \vspace{-2em}
   \caption{\textbf{S-MPI formulation and rendering.}
   Different from standard MPI with fronto-parallel image planes (a), S-MPI contains a set of adaptively posed planes (b). Each plane is represented by its normal $\mathbf{n}$ and offset $d$ to the optical center $\mathbf{o}$. (c) The rendering order of the noted pixel is given by [\textcircled{\raisebox{-0.9pt}{1}}, \textcircled{\raisebox{-0.9pt}{2}}, \textcircled{\raisebox{-0.9pt}{3}}], which follows the depth descending order, $[D_1, D_3, D_2]$. The depth value $D_i$ is obtained by backprojecting the pixel to plane $\pi_i$.
   }
   \label{fig:render}
   \vspace{-1em}
\end{figure}

\subsection{Rendering Formulation}
\label{sec:mpi-render-form}
\paragraph{S-MPI rendering.} 
Unlike the standard MPI having a global back-to-front rendering order of planes for each pixel due to the fronto-parallel planes, our S-MPI has different rendering orders for pixels because planes could intersect with each other, as shown in \cref{fig:render}. 
Each pixel has its own rendering order. 
First, we calculate depth values $D$ for each pixel on each plane. Let $\mathbf{K}$ denote the camera intrinsic. 
We backproject the 2D pixel $\mathbf{q}=(u, v, 1)$ to the 3D plane $\mathbf{\pi}=(\mathbf{n}, d)$, and then get the equation $ \mathbf{n} \cdot (D\mathbf{K}^{-1}\mathbf{q}) = d$ from \cref{eq:plane_eq}. Thus, the depth value takes the form:

\vspace{-1em}
\begin{equation}
  D = \frac{d}{\boldsymbol{n \cdot K^{-1}q}}.
  \label{eq:plane_depth}
\end{equation}

\noindent In the second step, we rearrange the \RGBA images with the depth order for each pixel.
The S-MPIs for each pixel $\mathbf{q}$ are sorted in the depth descending (back-to-front) order,  $\boldsymbol{\sigma} (\mathbf{q}) = [\sigma_1^{\mathbf{q}}, \sigma_2^{\mathbf{q}}, ..., \sigma_N^{\mathbf{q}} ] $, where $N=N_p + N_n$ is the total number of planar and non-planar proxies
and $\sigma_i^{\mathbf{q}}$ is the order index used in rendering for pixel $\mathbf{q}$ on the $i$-th proxy.
Then, we obtain the rearranged \RGBA images $[(C'_1, A'_1), (C'_2,A'_2), ..., (C'_N, A'_N)]$, where the \RGBA values of each pixel are given by $C'_i(\mathbf{q}) = C_{\sigma_i^{\mathbf{q}}}(\mathbf{q})$ and $A'_i(\mathbf{q}) = A_{\sigma_i^{\mathbf{q}}}(\mathbf{q})$.
Finally, we apply the standard alpha composition \cite{zhou2018stereo} with the new order to render the image $I$:
\vspace{-0.5em}
\begin{equation}
\vspace{-0.5em}
  I = \sum^{N}_{i=1}(C'_iA'_i\prod^{N}_{j=i+1}(1-A'_j)).
  \label{eq:alpha_compo}
\end{equation}

\noindent Similarly, a smooth depth map can be rendered by leveraging alpha to blend depth maps softly. First, we use \cref{eq:plane_depth} to get a depth map $\mathcal{D}_i$ for each plane. Then, we obtain the rearranged depth maps $[\mathcal{D}'_1, \mathcal{D}'_2, ..., \mathcal{D}'_N]$ with the new order $\boldsymbol{\sigma} (\mathbf{q})$. Finally, the rendered depth map $\mathcal{D_I}$ can be produced by $\mathcal{D_I} = \sum^{N}_{i=1}(\mathcal{D}'_iA'_i\prod^{N}_{j=i+1}(1-A'_i))$.

\begin{figure*}[t]
  \centering
  \includegraphics[width=0.9\linewidth]{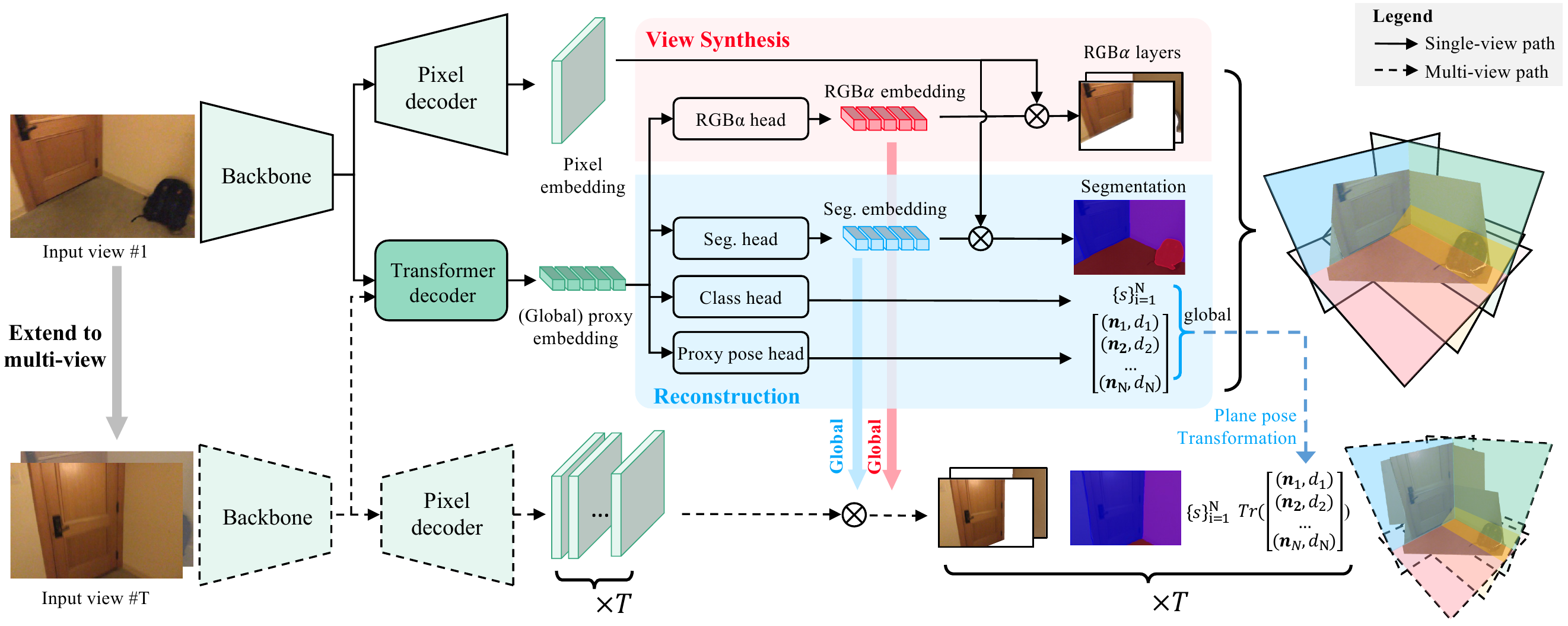}
  \vspace{-1.2em}
   \caption{\textbf{Structural MPI Transformer.} We use a backbone to extract features for $T$ ($=1$ or $>1$) images of different views and a pixel decoder extracts per-pixel embeddings. Then, a transformer decoder attends to multi-scale image features and produces $N$ (global) proxy embeddings. The proxy embeddings generate global class predictions and global plane poses which are then transformed to different view coordinate frames, ensuring plane pose alignment. Also, \RGBA embeddings and segmentation embeddings are generated globally. They are incorporated with pixel embeddings to generate $N \times T$ segmentation masks and $N \times T$ \RGBA layers with dynamic convolution layers.
   }
   \vspace{-1em}
   \label{fig:pipeline}
\end{figure*}
\vspace{-1em}
\paragraph{Rendering novel views.} To render a novel view image with S-MPI, we first transform the plane parameters from the source view. Given a transformation matrix $\boldsymbol{H} \in \mathbb{R}^{4 \times 4}$, a 3D point $\mathbf{x}'$ in the source view can be transferred to $\boldsymbol{H}\mathbf{x}'$. By \cref{eq:plane_eq}, 
We get the plane parameters in the target viewpoint as $\mathbf{\pi_t}=(\boldsymbol{H}^{-1})^T\mathbf{\pi_s}$. 
Then, we warp each plane of \RGBA image to target views by sampling from the source view with inverse homography \cite{hartley2003multiple}:
\vspace{-0.8em}
\begin{equation}
\vspace{-0.5em}
  [u_s, v_s, 1]^T = \boldsymbol{K(R}-\frac{\boldsymbol{tn}^T}{d})\boldsymbol{K}^{-1}[u_t,v_t,1]^T,
  \label{eq:homography}
\end{equation}

\noindent where $\boldsymbol{R,t}$ are rotation and translation decomposed from $\boldsymbol{H}$. Given transformed plane parameters and \RGBA images in the novel view, our rendering process can be applied to get the target RGB image.

%% file: Tex/4-method.tex
\section{Structural Multiplane Image Transformer}

Given $T$ views of images with known camera poses, our goal is to construct the S-MPI representation $\mathcal{P}_s$ for novel view synthesis and planar reconstruction. 
Most MPI construction methods \cite{zhou2018stereo, mine2021} assume that a pre-defined number of paralleled planes are preset. We propose a transformer-based model to predict S-MPI with an appropriate number of posed planes that faithfully approximate the scene, as well as \RGBA layers for further view synthesis. 
Our method is inspired by previous neural planar reconstruction methods \cite{liu2018planenet,tan2021planetr} that formulate plane detection as instance segmentation. Instead of neglecting non-planar regions or treating non-planar regions as a single layer, we differentiate non-planar instances according to their depth range and unify planar and non-planar detection in the same pipeline. 

Specifically, our model aims to predict proxies $\{(s_i, \mathbf{n}_i, d_i, M_i, C_i, A_i)\}_{i=1}^{N_p + N_n}$, where $s_i \in \{ \mathrm{planar}, \mathrm{nonplanar}\}$ denotes the structure class, $M_i$ is the visible projected mask of the plane in the current view. For multi-view input, we predict plane parameters $(\mathbf{n}_i, d_i)$ in a global coordinate frame for alignment.

\subsection{Single-view Network}
\label{sec:method-single-view}

\paragraph{Network design.} 
Our model is built upon a universal image segmentation network \cite{cheng2021maskformer, cheng2021mask2former}, which contains a pixel branch for pixel-level decoding and an instance branch representing instance-level embeddings. We fully utilize the two-branch architecture and propose a new one as illustrated in \cref{fig:pipeline}. 
First, the backbone extracts visual features and the pixel decoder upsamples features and generates high-resolution per-pixel embeddings. The transformer decoder operates on image features to generate $N$ proxy embeddings at the instance level.
Such proxy embeddings are used as queries for various downstream heads. Specifically, per-instance structure classification $\{s_i\}_{i=1}^{N}$ and plane parameter estimation $ \{ (\mathbf{n}_i, d_i)_{i=1}^{N} \}$ are generated from per-instance embeddings directly with linear layer heads. The segmentation and \RGBA predictions are generated by incorporating per-instance embeddings and high-resolution pixel embeddings with dynamic convolution \cite{cheng2021mask2former}. Finally, our S-MPI is created by combining predicted plane parameters and \RGBA layers of $N_p + N_n$ high-confidence instances.

\vspace{-1em}

\paragraph{Loss function.} To train our model, we jointly optimize view synthesis and planar estimation. 
For view synthesis, we render our predicted S-MPI in a novel view and employ $\mathcal{L}_{rgb}$, a combination of RGB L1 loss and SSIM loss between the rendered image and a ground truth image.
For the planar estimation, we employ $\mathcal{L}_{ce}$, a cross-entropy loss to distinguish planar and non-planar regions, $\mathcal{L}_{seg}$, a combination of focal loss \cite{lin2017focal} and dice loss \cite{milletari2016v} for segmentation, and $\mathcal{L}_{pln}$, an L1 loss for plane parameters estimation. 
Finally, we generate a depth map prediction by alpha compositing depth maps of each estimated plane by \cref{eq:plane_depth} and \cref{eq:alpha_compo} and employ $\mathcal{L}_{depth}$, a scale-invariant loss \cite{eigen2014depth} between the predicted and a ground truth depth map. Our total loss is:
\vspace{-0.8em}
\begin{equation}
\begin{aligned}
    \mathcal{L} &= \mathcal{L}_{view\_syn} + \mathcal{L}_{reconstruction}\\
    &= \mathcal{L}_{rgb} + (\alpha \mathcal{L}_{ce} + \beta \mathcal{L}_{seg} + \delta \mathcal{L}_{pln} + \mathcal{L}_{depth}).
\end{aligned}
\label{eq:total_loss}
\end{equation}

\subsection{Multi-view Network}
\label{sec:method-multi-view}

\begin{figure}[t]
  \centering
  \includegraphics[width=1.0\linewidth]{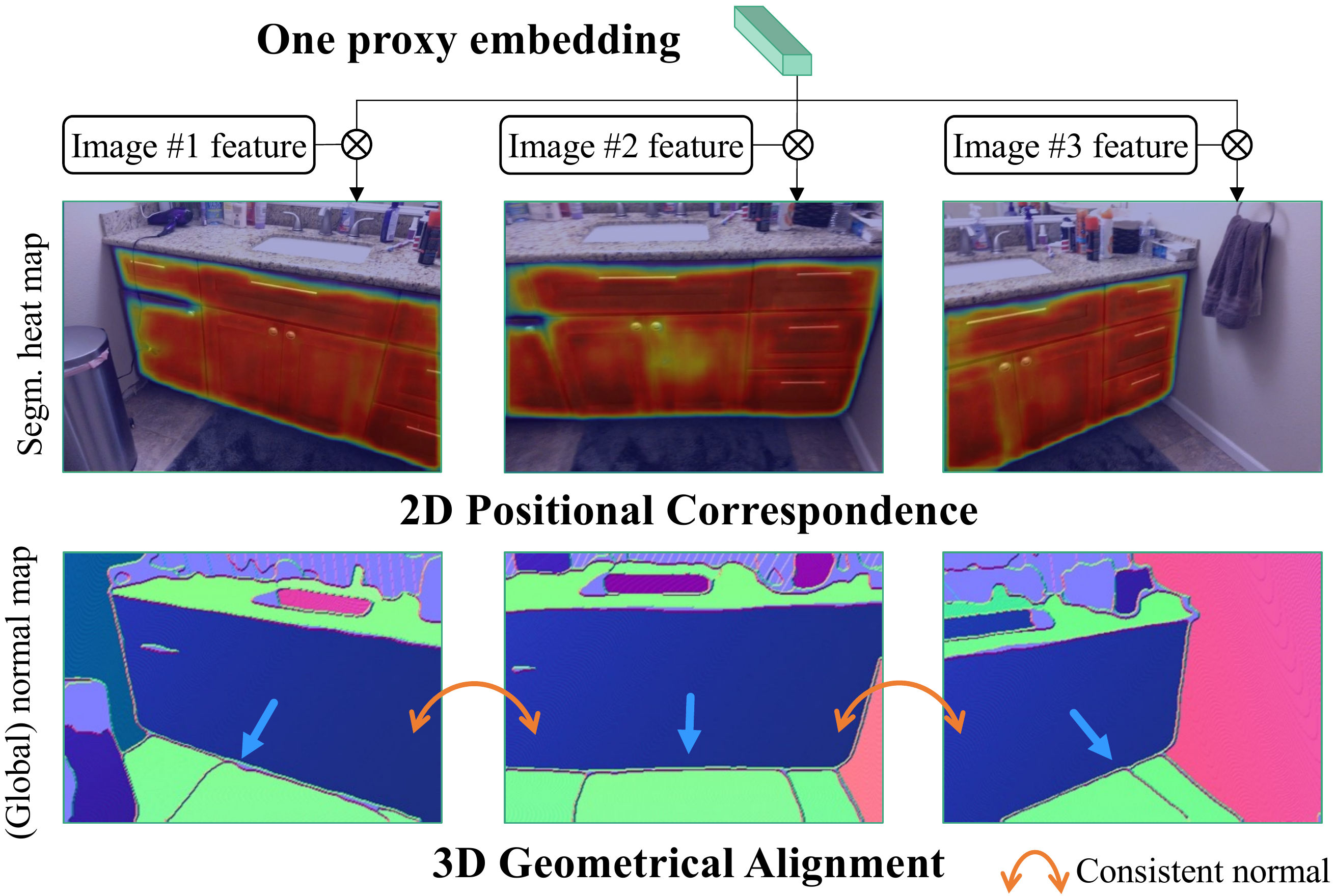}
  \vspace{-2em}
   \caption{Both 2D positional and 3D geometry information are encoded in our instance embedding. By dot production with image features, the corresponding instances in different views can be visualized. In the normal map, identical plane normals in the global space are painted with the same color.
   }
   \label{fig:corr}
   \vspace{-1.5em}
\end{figure}

Multi-view input can enlarge the range of view synthesis, yet it also introduces new challenges.
As to processing each view independently, single-view methods generating planar reconstructions are hardly aligned in the 3D space. Our goal is to deliver multi-view consistent planar reconstruction for better view synthesis.

\vspace{-1.5em}

\paragraph{Network design.} Inspired by the application of transformer in video instance segmentation \cite{cheng2021mask2formerv}, as the proxy embeddings represent instance-level information of proxies in the current view in \cref{sec:method-single-view}, we extend them to the global scene extent. Thus, the proxy embeddings across views are shared, representing all proxies in a global space. As is shown in \cref{fig:pipeline}, given $T$ images, we first generate $T$ pixel embeddings with the shared backbone and pixel decoder. Then, the global proxy embeddings are utilized to generate proxies in the same way as the single-view network with the multi-view alignment.

\vspace{-1.5em}

\paragraph{Multi-view alignment.} As illustrated in \cref{fig:pipeline}, the global proxy embedding is shared across $T$ views, as well as its outputs from the four heads.
Thus, the predicted structure classes $\{s\}_{i=1}^N$ and poses $[(\mathbf{n}_1,d_1),(\mathbf{n}_2,d_2),...,(\mathbf{n}_N,d_N)]$ of all planes are the same in $T$ views. We consider these shared plane parameters to be a set of global plane poses in the full extent of the scene. Given camera poses, we then transform the global plane poses to the $T$ views so that corresponding instances in each view are naturally aligned geometrically. Also, \RGBA embeddings and segmentation embeddings are shared globally. 
By using these global proxy-level embeddings to query $T$ pixel embeddings, 
each global proxy decodes out its corresponding segment mask and \RGBA layer in each of the $T$ views. Then, we employ the supervision in \cref{sec:method-single-view} for $T$ views. In this way, the global proxy embeddings are learned with the ensembled supervision from all views with proxy-wise consistency, which can progressively cover regions to the full extent of scenes and refine predictions to obey the alignment.
As illustrated in \cref{fig:corr}, the class activation maps and globally aligned normal maps from an example proxy embedding appear to be consistent.

\vspace{-1.5em}

\paragraph{Image merging.}
The final image generation is by merging generated images from multiple views.
For each input view, we render an image in the target view and save the alpha weights used in the alpha composition process. Then, we follow \cite{mildenhall2019llff} to fuse the $T$ rendered images, where the alpha weights are used as confidence maps to blend images. Areas with low confidence in one view that are occluded or beyond the canvas will be overlaid by content with high confidence in other views.

%% file: Tex/5-experiment.tex
\section{Experiments}

We demonstrate the effectiveness of our S-MPI by showing its state-of-the-art performance on both view synthesis and reconstruction from single-view and multi-view settings. 
Then we ablate the design of S-MPI confirming that the improvements stem from specific components.

\vspace{-1.5em}
\paragraph{Datasets.}
We use the image-based dataset, NYUv2 \cite{silberman2012indoor}, for single-view reconstruction. 
The video-based dataset ScanNet \cite{dai2017scannet} is adopted for single-view and multi-view reconstruction and view synthesis.

\input{Tex/tabs/dep_sv.tex}
\input{Tex/tabs/syn_sv.tex}

\vspace{-0.5em}

\subsection{Implementation}

\vspace{-0.5em}
\paragraph{Data preparation.}
For the ScanNet dataset, we use the same ground truth labels of plane instance masks and plane parameters as \cite{liu2018planenet}.
For the non-planar regions, we simply sample S-MPIs according to the depth range of the scene uniformly and produce a set of masks of depth-wise segmentation.
Then, we exclude planar masks as non-planar mask labels. 
It is worth noting that sophisticated depth division \cite{han2022single,luvizon2021adaptive} can be easily applied in our pipeline. We do not discuss the depth division for non-planar regions as it is not our main focus.
The images are resized to $256 \times384$ for training and evaluation. 

\vspace{-1em}
\begin{figure}[b]
\vspace{-1em}
  \centering
  \includegraphics[width=0.75\linewidth]{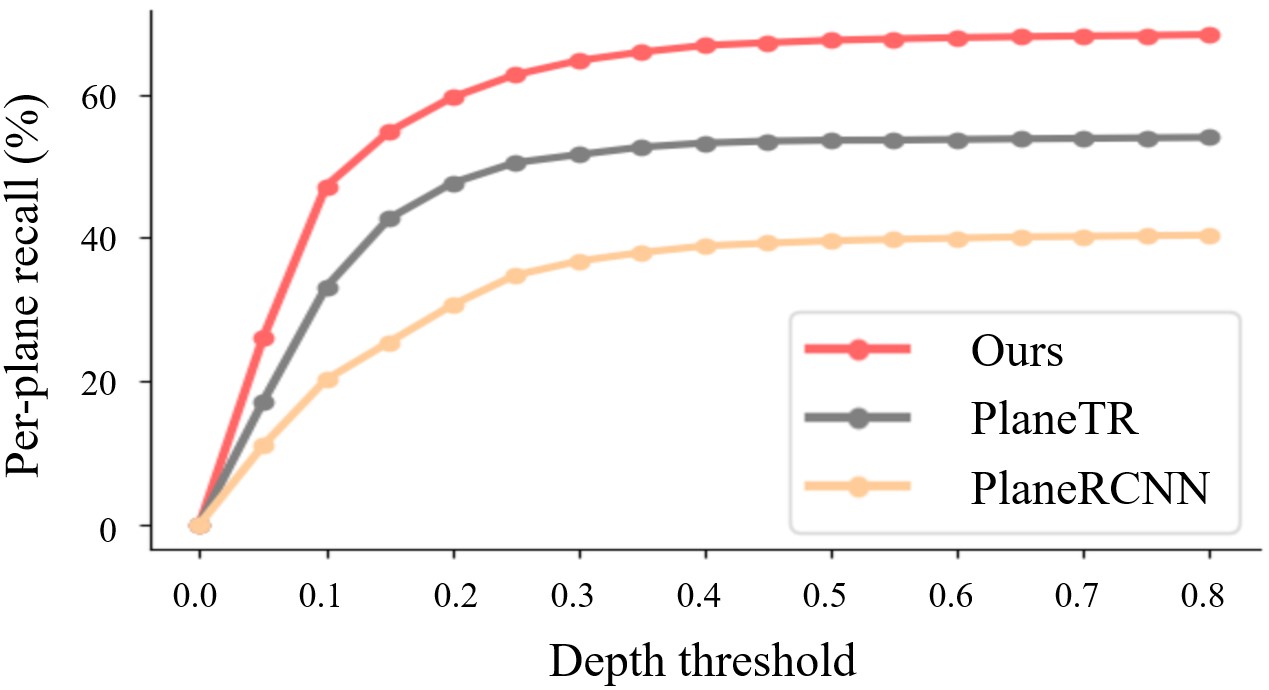}
  \vspace{-1em}
   \caption{Single-view planar estimation results on ScanNet \cite{dai2017scannet} by measuring the plane recall with a fixed Intersection over Union threshold 0.5 and a varying depth error threshold (from 0 to 0.8m).}
   \label{fig:exp-recall}

\end{figure}

\paragraph{Implementation details.}
We use ResNet50 \cite{he2016deep} as the backbone. The Adam Optimizer is used with an initial learning rate of 0.0001 and a weight decay of 0.05 in our training. In \cref{eq:total_loss}, $\alpha=2, \beta=5, \delta=5$. The model is trained on 4 NVIDIA-V100 GPUs for a total of 100k steps.

\vspace{-1em}
\paragraph{Training.}
Before our main training, we add a bootstrap training phase for 50k steps, where we initialize the alpha prediction with the segmentation by turning off $\mathcal{L}_{rgb}$ and enforcing a segmentation loss on the alpha channel. To train the model with multi-view input, we set $T=2$ and the trained model can be applied for $T \geq 1$. We cancel $\mathcal{L}_{pln}$ for non-planar regions in the second input view.

\input{Tex/tabs/rec_sv.tex}

\input{Tex/tabs/ablation}

\input{Tex/tabs/rec_mv.tex}

\vspace{-0.5em}

\subsection{Single-view Evaluation}

\vspace{-0.5em}

\paragraph{Reconstruction.} For planar reconstruction, we follow PlaneRCNN \cite{liu2019planercnn} and PlaneTR \cite{tan2021planetr} to evaluate planar estimation metrics (per-plane recall) in  \cref{fig:exp-recall}, and segmentation metrics (Variation of Information, Rand Index, and Segmentation Covering) in \cref{tab:sin-planeseg}. For depth estimation, we compare with planar reconstruction methods \cite{liu2019planercnn, tan2021planetr} and MPI-based view synthesis methods \cite{tucker2020single, mine2021} in \cref{tab:sin-depth}. All the methods are trained on ScanNet \cite{dai2017scannet} and evaluated on NYUv2 \cite{silberman2012indoor} and ScanNet.

\vspace{-1.5em}
\paragraph{View synthesis.} We compare with the standard MPI-based methods MPI \cite{tucker2020single} and MINE \cite{mine2021} in \cref{tab:sin-nvs}. All the methods are trained on ScanNet \cite{dai2017scannet} with losses including $ \mathcal{L}_{rgb}$ and $\mathcal{L}_{depth}$ in \cref{eq:total_loss} given GT images and depth maps of source and target views. Note that we add the $\mathcal{L}_{depth}$ when training standard MPI methods for a fair comparison.

The results show that our method achieves better reconstruction and view synthesis performance. The reason is that our method benefits from proxy-level plane parameter prediction with transformer and alpha blended depth map generation in both planar and non-planar regions. Also, because of our adaptive planar structure and accurate scene reconstruction, our method outperforms standard MPI.

\subsection{Multi-view Evaluation}

\vspace{-0.5em}

\paragraph{Reconstruction.} 
We compare with PlaneMVS \cite{liu2022planemvs} on plane detection and depth reconstruction accuracy in \cref{tab:mv-plane}. Plane detection is measured by Average Precision with IoU 0.5 and a varying depth error (from 0.2 to 0.9m). Both methods are trained with paired images input from ScanNet. We prepare the test data following PlaneMVS consisting of image pairs with a translation from 0.05 to 0.15m. The results show the advantages of our global proxy embedding which incorporates multi-view information contributing to a global geometrical representation.

\vspace{-1.5em}

\begin{figure}[b]
\vspace{-1em}
  \centering
  \includegraphics[width=1.0\linewidth]{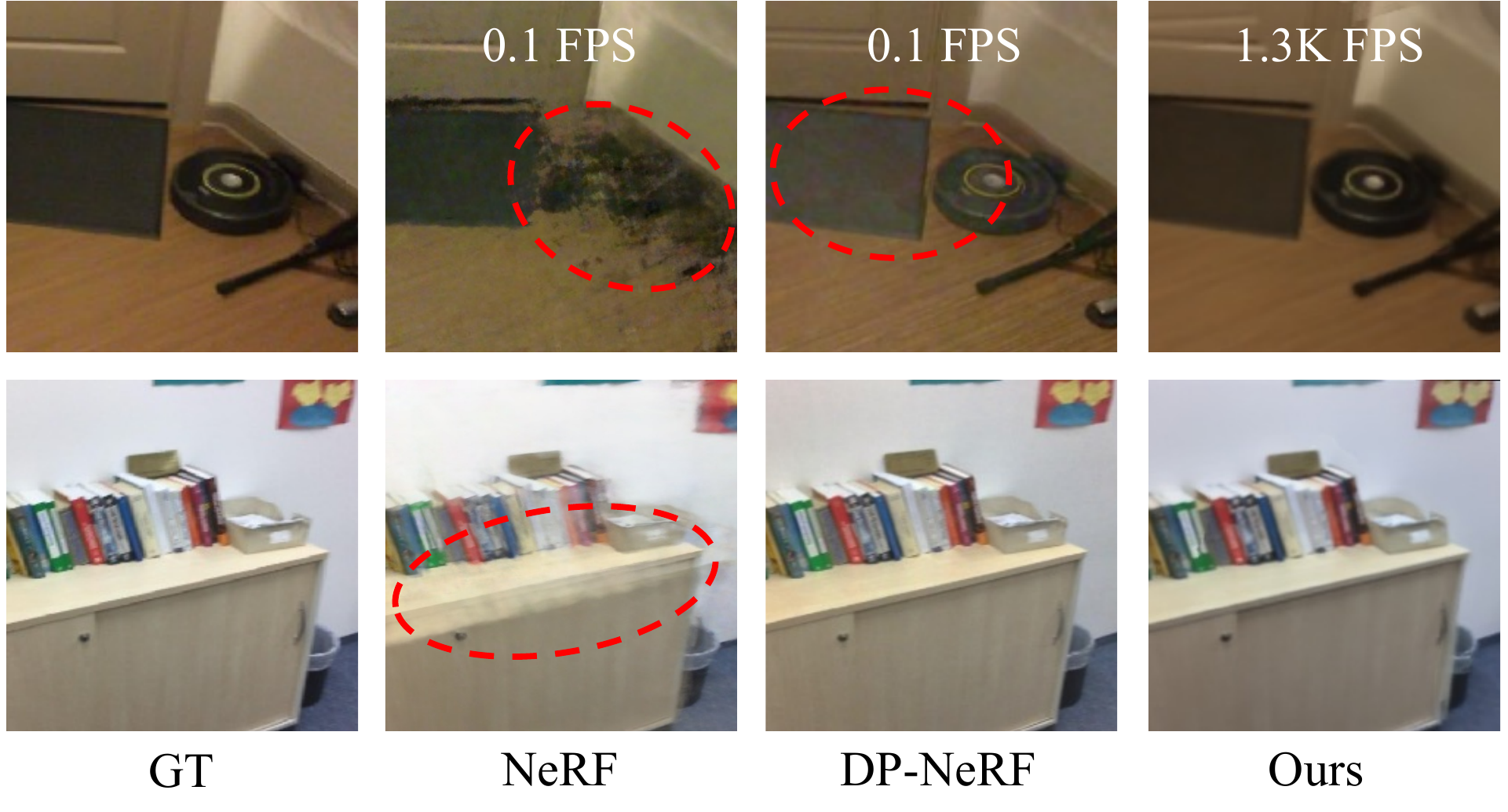}
  \vspace{-2em}
  \caption{Compared to NeRF-based methods (NeRF \cite{mildenhall2020nerf}, DP-NeRF \cite{roessle2022dense}), we achieve better  results with less noise and faster rendering speed, while ours do not need per-scene training.}
  \label{fig:exp-nerf}
\end{figure}

\paragraph{View Synthesis.} We follow the data settings in DP-NeRF \cite{roessle2022dense} where they sample images sparsely in selected scenes in ScanNet and generate dense depth maps for additional supervision. The average view gap between two views is 68 frames which is much more challenging than our paired training data (20 frames). Our training scenes have no overlap with their test set, while NeRF-based methods \cite{mildenhall2020nerf, deng2022depth, roessle2022dense, wei2021nerfingmvs} are trained on the test scenes. 
For each test image, we select the nearest two views as our input. For a fair comparison, we additionally train DP-NeRF on the two nearest images and the results are noted with ``(2)''. To test MINE \cite{mine2021}, we follow \cite{mildenhall2019llff} to merge two MPI-generated images.
The results in \cref{tab:mul-nerf} show that our method outperforms MPI-based methods and achieves comparable results to NeRF-based methods, while no training in the target scene is performed. Qualitative comparisons are shown in \cref{fig:exp-nerf}.
We also compare the rendering speed with NeRF based method in the last column in \cref{tab:mul-nerf}.
Our method inherits the speed advantage of MPI-based methods. The speed of rendering depends on the number of plane layers, while this number is adaptive to different scenes in our method. The average number of planes in ScanNet is about 12.5. Our method is slower than MINE since we need additional plane intersection checking, but it is much faster than NeRF-based methods.

\input{Tex/tabs/syn_mv.tex}

\subsection{Ablation Study}
We conduct the ablation study on ScanNet \cite{dai2017scannet}. Our multi-view model is trained with input image pairs with a gap of $<20$ frames and the view synthesis target image is $<30$ frames away from the middle of two input images.

\noindent \textbf{Multi-view consistency.} \cref{tab:ablation-query-number} shows that with our global proxy embedding strategy, synthesized images are better aligned than processing multi-view images one by one. 

\noindent \textbf{Difference of input image pairs.} \cref{tab:abla-pair-diff} shows that our method benefits from the consistent multi-view geometry supervision and produces more accurate planar reconstruction than given two identical images (gap = 0). However, the performance drops inevitably if the view gap is too large.

\noindent \textbf{Percentage of plane area.} We divide our test data according to the planar area ratio. \cref{tab:abla-planar-ratio} shows that our method performs better in scenes with more planar coverage. When there are few planar regions, our method degrades to standard MPI with adaptive plane numbers and our performance is compatible with MINE \cite{mine2021}. To further validate it, we force $N_p$ in \cref{sec:mpi-geo-form} to be 0 in the groundtruth to regard all regions as non-planar (0\%* in \cref{tab:abla-planar-ratio}) to make comparison.

\noindent \textbf{Query number.} \cref{tab:ablation-query-number} shows a marginal performance improvement as the query number increases in our transformer, as the numbers of planes are usually not large in man-made scenes.

\begin{figure*}[t]
  \centering
  \includegraphics[width=1.0\linewidth]{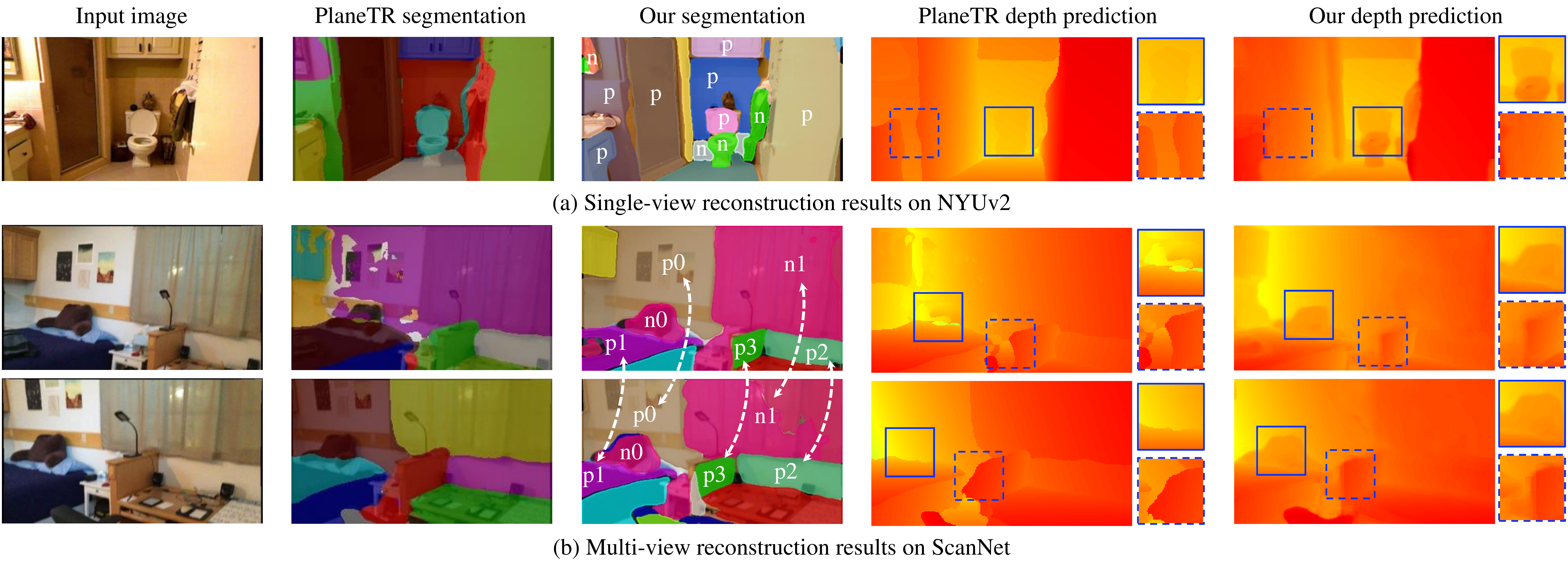}
  \vspace{-2em}
  \caption{\textbf{Planar reconstruction results on (single-view) NYUv2 \cite{silberman2012indoor} and (multi-view) ScanNet \cite{dai2017scannet}}. Compared with PlaneTR \cite{tan2021planetr}, our method can uniformly predict planar (p) and non-planar (n) instances and multi-view consistent segmentation and geometry prediction with matched instances.}
  \label{fig:exp-recon}
\end{figure*}

\begin{figure*}
\vspace{-0.5em}
  \centering
  \includegraphics[width=1.0\linewidth]{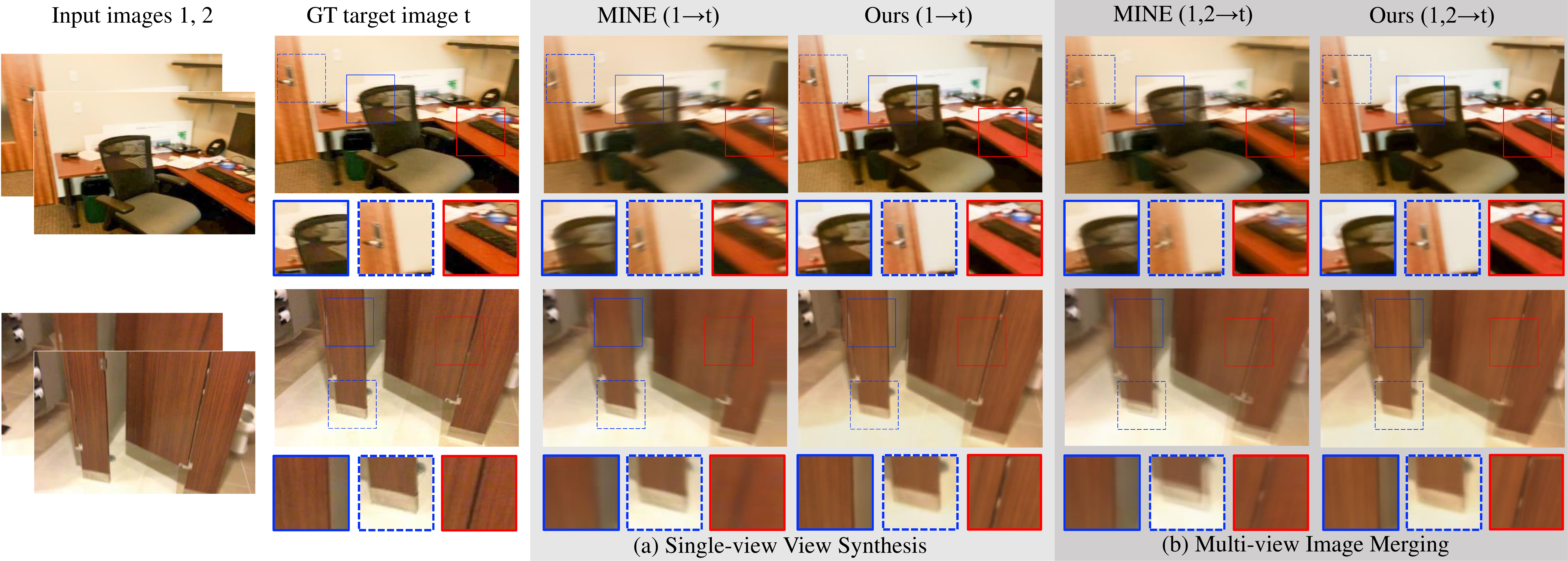}
  \caption{\textbf{View synthesis results on ScanNet \cite{dai2017scannet}}. We show single-view results (1$\to$t) and multi-view results (1,2$\to$t) in comparison with MINE \cite{mine2021}. The multi-view results of MINE are generated by image merging following \cite{mildenhall2019llff}. }
  \vspace{-1em}
  \label{fig:exp-viewsyn}
\end{figure*}

%% file: Tex/tabs/dep_sv.tex
\begin{table*}[t]
\centering
\caption{Single-view depth estimation results on NYUv2 \cite{silberman2012indoor} and ScanNet \cite{dai2017scannet}. The target view depth maps are acquired by novel view synthesis methods with source view images input. All the methods are trained on ScanNet and our method achieves higher accuracy than planar reconstruction methods\cite{tan2021planetr,liu2019planercnn} and MPI-based methods \cite{tucker2020single,mine2021}.}
\vspace{-0.8em}
\scalebox{0.8}{
\begin{tabular}{c|ccccc|ccc|ccc} 
\hline
 & \multicolumn{5}{c|}{NYUv2} & \multicolumn{3}{c|}{ScanNet (src view)} & \multicolumn{3}{c}{ScanNet (tgt view)} \\
 & MPI\cite{tucker2020single} & MINE\cite{mine2021} & PlaneTR\cite{tan2021planetr} & Plane\small{RCNN}\cite{liu2019planercnn} & Ours & MPI & MINE & Ours & MPI & MINE & Ours \\ 
\hline
rel↓ & 0.241 & 0.215 & 0.265 & 0.164 & \textbf{0.155} & 0.149 & 0.128 & \textbf{0.082} & 0.155 & 0.147 & \textbf{0.101} \\
log↓ & 0.097 & 0.094 & 0.084 & 0.074 & \textbf{0.065} & 0.056 & 0.051 & \textbf{0.030} & 0.074 & 0.069 & \textbf{0.053} \\
rmse↓ & 0.775 & 0.754 & 0.686 & 0.644 & \textbf{0.530} & 0.308 & 0.288 & \textbf{0.214} & 0.320 & 0.303 & \textbf{0.251} \\
a1↑ & 0.665 & 0.683 & 0.734 & 0.753 & \textbf{0.779} & 0.858 & 0.914 & \textbf{0.964} & 0.842 & 0.887 & \textbf{0.946} \\
a2↑ & 0.874 & 0.894 & 0.913 & 0.931 & \textbf{0.956} & 0.965 & 0.970 & \textbf{0.985} & 0.943 & 0.959 & \textbf{0.979} \\
a3↑ & 0.950 & 0.952 & 0.961 & 0.982 & \textbf{0.990} & 0.984 & 0.989 & \textbf{0.996} & 0.976 & 0.978 & \textbf{0.982} \\
\hline
\end{tabular}
}
\label{tab:sin-depth}
\vspace{-0.8em}
\end{table*}

%% file: Tex/tabs/syn_sv.tex
\begin{table*}[t]
\centering
\caption{Single-view view synthesis results on ScanNet \cite{dai2017scannet}. $n$ is the number of frames between source views and target views. The numbers (32,64) for MPI and MINE are pre-set numbers of MPI layers.
}
\vspace{-0.8em}
\scalebox{0.8}{
\begin{tabular}{c|ccc|ccc|ccc} 
\hline
 & \multicolumn{3}{c|}{LPIPS$\downarrow$} & \multicolumn{3}{c|}{SSIM$\uparrow$} & \multicolumn{3}{c}{PSNR$\uparrow$} \\
 & $n<15$ & $n<30$ & $n<45$ & $n<15$ & $n<30$ & $n<45$ & $n<15$ & $n<30$ & $n<45$ \\ 
\hline
MPI-32  \cite{tucker2020single} & 0.2632 & 0.3365 & 0.4331 & 0.8081 & 0.7567 & 0.7328 & 22.737 & 20.405 & 19.358 \\
MINE-32 \cite{mine2021} & 0.2498 & 0.3064 & 0.3502 & 0.8220 & 0.7859 & 0.7487 & 22.514 & 20.803 & 19.517 \\
MINE-64 \cite{mine2021} & 0.2412 & 0.2845 & 0.3454 & 0.8296 & 0.7936 & 0.7561 & 22.862 & 21.058 & 19.705 \\
Ours & \textbf{\textbf{0.1853}} & \textbf{\textbf{0.1914}} & \textbf{0.1984} & \textbf{\textbf{0.8412}} & \textbf{\textbf{0.8285}} & \textbf{0.8218} & \textbf{\textbf{25.022}} & \textbf{\textbf{23.604}} & \textbf{23.375} \\
\hline
\end{tabular}
}
\label{tab:sin-nvs}
\vspace{-1.5em}
\end{table*}

%% file: Tex/tabs/rec_sv.tex
\begin{table}[b]
\centering
\vspace{-1.5em}
\caption{Single-view planar segmentation results on ScanNet \cite{dai2017scannet}. 
``VI'', ``RI'' and ``SC'' denote Variation of Information, Rand Index, and Segmentation Covering, respectively.
}
\vspace{-0.5em}
\scalebox{0.8}{
\begin{tabular}{l|rrr} 
\hline
 & \multicolumn{1}{l}{VI↓} & \multicolumn{1}{l}{RI↑} & \multicolumn{1}{l}{SC↑} \\ 
\hline
PlaneNet\cite{liu2018planenet} & 2.142 & 0.797 & 0.692 \\
PlaneRCNN\cite{liu2019planercnn} & 1.809 & 0.880 & 0.810 \\
PlaneTR\cite{tan2021planetr} & 0.898 & 0.924 & 0.811 \\
Ours & \textbf{0.791} & \textbf{0.947} & \textbf{0.851} \\
\hline
\end{tabular}
}
\label{tab:sin-planeseg}
\end{table}

%% file: Tex/tabs/ablation.tex
\begin{table*}
\centering
\makebox[0pt][c]{\parbox{1\textwidth}{%
    \begin{minipage}[t]{0.32\hsize}\centering
        \caption{Ablation study on multi-view consistency (MVC) and query number (the maximum number of \RGBA layers).}
        \vspace{-0.4em}
        \scalebox{0.78}{
        \begin{tabular}{c|c|ccc} 
        \hline
        \# query & MVC & LPIPS↓ & SSIM↑ & PSNR↑ \\ 
        \hline
        25 & $\checkmark$ & 0.199 & 0.803 & 22.97 \\
        50 & $\checkmark$ & 0.197 & 0.805 & 23.06 \\
        100 &  & 0.208 & 0.749 & 22.65 \\
        100 & $\checkmark$ & \textbf{0.197} & \textbf{0.813} & \textbf{23.11} \\
        \hline
        \end{tabular}
        }
        \label{tab:ablation-query-number}
    \end{minipage}
    \hfill
    \begin{minipage}[t]{0.32\hsize}\centering
        \caption{Ablation study on the difference (\# frames gap) of input image pairs. Plane recalls with depth (0.1m,0.6m) and normal (5$^\circ$,30$^\circ$) error are evaluated.}
        \vspace{-0.6em}
        \scalebox{0.78}{
        \begin{tabular}{c|cccc} 
        \hline
        \# gap & RC$^{0.1m}$ & RC$^{0.6m}$ & RC$^{5^\circ}$ & RC$^{30^\circ}$ \\ 
        \hline
        0 & 52.61 & 76.43 & \textbf{48.02} & 74.76 \\
        20 & \textbf{52.95} & \textbf{77.16} & 47.17 & \textbf{75.19} \\
        40 & 46.43 & 75.47 & 43.02 & 73.33 \\
        \hline
        \end{tabular}
        }
        \label{tab:abla-pair-diff}
    \end{minipage}
    \hfill
    \begin{minipage}[t]{0.32\hsize}\centering
        \caption{Ablation study on the percentage of plane area in the whole image.}
        \vspace{-0.3em}
        \scalebox{0.78}{
        \begin{tabular}{c|ccc} 
        \hline
        planar area & LPIPS↓ & SSIM↑ & PSNR↑ \\ 
        \hline
        0\%\textasciitilde{}40\% & 0.208 & 0.795 & 22.97 \\
        40\%\textasciitilde{}60\% & 0.196 & 0.807 & 23.61 \\
        60\%\textasciitilde{}80\% & 0.185 & 0.812 & 24.03 \\
        80\%\textasciitilde{}100\% & \textbf{0.179} & \textbf{0.841} & \textbf{24.52} \\
        \hline
        0\%* & 0.212 & 0.796 & 21.10 \\
        \hline
        \end{tabular}
        }
        \label{tab:abla-planar-ratio}
    \end{minipage}
}}
\vspace{-1em}
\end{table*}

%% file: Tex/tabs/rec_mv.tex
\begin{table}[t]
\centering
\caption{Multi-view reconstruction results on ScanNet \cite{dai2017scannet}. We evaluate depth accuracy and planar detection results in comparison with PlaneMVS \cite{liu2022planemvs}. Both methods take image pairs as input. (Translation: 0.05$\sim$0.15m)}
\vspace{-0.5em}
\scalebox{0.8}{
\begin{tabular}{c|cc|c|cc} 
\hline
 & \small{Pln-MVS} & Ours &  & \small{Pln-MVS} & Ours \\ 
\hline
rel↓ & 0.088 & \textbf{0.079} & \small{AP$^{0.2m}$}↑ & 0.456 & \textbf{0.579} \\
rmse↓ & \textbf{0.186} & 0.205 & \small{AP$^{0.4m}$}↑ & 0.540 & \textbf{0.649} \\
a1↑ & 0.926 & \textbf{0.946} & \small{AP$^{0.6m}$}↑ & 0.559 & \textbf{0.704} \\
a2↑ & 0.988 & \textbf{0.989} & \small{AP$^{0.9m}$}↑ & 0.562 & \textbf{0.716} \\
\hline
\end{tabular}
}
\label{tab:mv-plane}
\vspace{-1.5em}
\end{table}

%% file: Tex/tabs/syn_mv.tex
\begin{table}[t]
\centering
\caption{Quantitative view synthesis results with multi-view inputs on ScanNet. 
``($x$)'' indicates using $x$ neighboring images of the test image in training for NeRF based methods, and $x$ input views in inference for MINE \cite{mine2021} and ours. 
```TPS'' is short for training per scene. The bold number indicate the best performance and the \textcolor{blue}{\underline{blue underlying}} ones indicate the best performance without TPS.}
\vspace{-0.5em}
\scalebox{0.8}{
\begin{tabular}{c|c|ccc|c} 
\hline
 & TPS & {LPIPS↓} & {SSIM↑} & {PSNR↑} & FPS \\
\hline
NeRF \cite{mildenhall2020nerf} (18) & \checkmark & 0.398 & 0.670 & 19.03 & 0.1 \\
DS-NeRF \cite{deng2022depth} (18) & \checkmark & 0.344 & 0.713 & 20.85 & - \\
NerfingMVS \cite{wei2021nerfingmvs} (18) & \checkmark & 0.502 & 0.626 & 16.29 & -\\
DP-NeRF (18) \cite{roessle2022dense} & \checkmark & 0.294 & \textbf{0.737} & \textbf{20.96} & 0.1 \\
\hline
DP-NeRF \cite{roessle2022dense} (2) & \checkmark  & 0.324 & 0.712 & 20.49 & 0.1 \\ 
MINE \cite{mine2021} (2) &  & 0.359 & 0.635 & 16.79 & \underline{\textbf{\textcolor{blue}{2.5K}}} \\
Ours (2) & & \underline{\textbf{\textcolor{blue}{0.267}}} & \underline{\textcolor{blue}{0.703}} & \underline{\textcolor{blue}{19.93}} & 1.3K\\
\hline
\end{tabular}
}
\label{tab:mul-nerf}
\vspace{-1.5em}
\end{table}


%% file: CameraReady.bbl
\begin{thebibliography}{10}\itemsep=-1pt

\bibitem{agarwala2022planes}
Samir Agarwala, Linyi Jin, Chris Rockwell, and David~F. Fouhey.
\newblock Planeformers: From sparse view planes to 3d reconstruction.
\newblock In {\em ECCV}, 2022.

\bibitem{chen2021mvsnerf}
Anpei Chen, Zexiang Xu, Fuqiang Zhao, Xiaoshuai Zhang, Fanbo Xiang, Jingyi Yu,
  and Hao Su.
\newblock Mvsnerf: Fast generalizable radiance field reconstruction from
  multi-view stereo.
\newblock In {\em Proceedings of the IEEE/CVF International Conference on
  Computer Vision}, pages 14124--14133, 2021.

\bibitem{cheng2021mask2formerv}
Bowen Cheng, Anwesa Choudhuri, Ishan Misra, Alexander Kirillov, Rohit Girdhar,
  and Alexander~G Schwing.
\newblock Mask2former for video instance segmentation.
\newblock {\em arXiv preprint arXiv:2112.10764}, 2021.

\bibitem{cheng2021mask2former}
Bowen Cheng, Ishan Misra, Alexander~G. Schwing, Alexander Kirillov, and Rohit
  Girdhar.
\newblock Masked-attention mask transformer for universal image segmentation.
\newblock 2022.

\bibitem{cheng2021maskformer}
Bowen Cheng, Alexander~G. Schwing, and Alexander Kirillov.
\newblock Per-pixel classification is not all you need for semantic
  segmentation.
\newblock 2021.

\bibitem{dai2017scannet}
Angela Dai, Angel~X. Chang, Manolis Savva, Maciej Halber, Thomas Funkhouser,
  and Matthias Nie{\ss}ner.
\newblock Scannet: Richly-annotated 3d reconstructions of indoor scenes.
\newblock In {\em Proc. Computer Vision and Pattern Recognition (CVPR), IEEE},
  2017.

\bibitem{deng2022depth}
Kangle Deng, Andrew Liu, Jun-Yan Zhu, and Deva Ramanan.
\newblock Depth-supervised nerf: Fewer views and faster training for free.
\newblock In {\em Proceedings of the IEEE/CVF Conference on Computer Vision and
  Pattern Recognition}, pages 12882--12891, 2022.

\bibitem{dhamo2019peeking}
Helisa Dhamo, Keisuke Tateno, Iro Laina, Nassir Navab, and Federico Tombari.
\newblock Peeking behind objects: Layered depth prediction from a single image.
\newblock {\em Pattern Recognition Letters}, 125:333--340, 2019.

\bibitem{eigen2014depth}
David Eigen, Christian Puhrsch, and Rob Fergus.
\newblock Depth map prediction from a single image using a multi-scale deep
  network.
\newblock {\em Advances in neural information processing systems}, 27, 2014.

\bibitem{flynn2019deepview}
John Flynn, Michael Broxton, Paul Debevec, Matthew DuVall, Graham Fyffe, Ryan
  Overbeck, Noah Snavely, and Richard Tucker.
\newblock Deepview: View synthesis with learned gradient descent.
\newblock In {\em Proceedings of the IEEE/CVF Conference on Computer Vision and
  Pattern Recognition}, pages 2367--2376, 2019.

\bibitem{gallup2007real}
David Gallup, Jan-Michael Frahm, Philippos Mordohai, Qingxiong Yang, and Marc
  Pollefeys.
\newblock Real-time plane-sweeping stereo with multiple sweeping directions.
\newblock In {\em 2007 IEEE Conference on Computer Vision and Pattern
  Recognition}, pages 1--8. IEEE, 2007.

\bibitem{gallup2010piecewise}
David Gallup, Jan-Michael Frahm, and Marc Pollefeys.
\newblock Piecewise planar and non-planar stereo for urban scene
  reconstruction.
\newblock In {\em 2010 IEEE computer society conference on computer vision and
  pattern recognition}, pages 1418--1425. IEEE, 2010.

\bibitem{han2022single}
Yuxuan Han, Ruicheng Wang, and Jiaolong Yang.
\newblock Single-view view synthesis in the wild with learned adaptive
  multiplane images.
\newblock In {\em ACM SIGGRAPH}, 2022.

\bibitem{hartley2003multiple}
Richard Hartley and Andrew Zisserman.
\newblock {\em Multiple view geometry in computer vision}.
\newblock Cambridge university press, 2003.

\bibitem{he2016deep}
Kaiming He, Xiangyu Zhang, Shaoqing Ren, and Jian Sun.
\newblock Deep residual learning for image recognition.
\newblock In {\em Proceedings of the IEEE conference on computer vision and
  pattern recognition}, pages 770--778, 2016.

\bibitem{hu2021worldsheet}
Ronghang Hu, Nikhila Ravi, Alexander~C Berg, and Deepak Pathak.
\newblock Worldsheet: Wrapping the world in a 3d sheet for view synthesis from
  a single image.
\newblock In {\em Proceedings of the IEEE/CVF International Conference on
  Computer Vision}, pages 12528--12537, 2021.

\bibitem{ikehata2015structured}
Satoshi Ikehata, Hang Yang, and Yasutaka Furukawa.
\newblock Structured indoor modeling.
\newblock In {\em Proceedings of the IEEE international conference on computer
  vision}, pages 1323--1331, 2015.

\bibitem{jin2021planar}
Linyi Jin, Shengyi Qian, Andrew Owens, and David~F Fouhey.
\newblock Planar surface reconstruction from sparse views.
\newblock In {\em Proceedings of the IEEE/CVF International Conference on
  Computer Vision}, pages 12991--13000, 2021.

\bibitem{johari2022geonerf}
Mohammad~Mahdi Johari, Yann Lepoittevin, and Fran{\c{c}}ois Fleuret.
\newblock Geonerf: Generalizing nerf with geometry priors.
\newblock In {\em Proceedings of the IEEE/CVF Conference on Computer Vision and
  Pattern Recognition}, pages 18365--18375, 2022.

\bibitem{mine2021}
Jiaxin Li, Zijian Feng, Qi She, Henghui Ding, Changhu Wang, and Gim~Hee Lee.
\newblock Mine: Towards continuous depth mpi with nerf for novel view
  synthesis.
\newblock In {\em ICCV}, 2021.

\bibitem{li2020synthesizing}
Qinbo Li and Nima~Khademi Kalantari.
\newblock Synthesizing light field from a single image with variable mpi and
  two network fusion.
\newblock {\em ACM Trans. Graph.}, 39(6):229--1, 2020.

\bibitem{lin2017focal}
Tsung-Yi Lin, Priya Goyal, Ross Girshick, Kaiming He, and Piotr Doll{\'a}r.
\newblock Focal loss for dense object detection.
\newblock In {\em Proceedings of the IEEE international conference on computer
  vision}, pages 2980--2988, 2017.

\bibitem{lin2022neurmips}
Zhi-Hao Lin, Wei-Chiu Ma, Hao-Yu Hsu, Yu-Chiang~Frank Wang, and Shenlong Wang.
\newblock Neurmips: Neural mixture of planar experts for view synthesis.
\newblock In {\em Proceedings of the IEEE/CVF Conference on Computer Vision and
  Pattern Recognition}, pages 15702--15712, 2022.

\bibitem{liu2019planercnn}
Chen Liu, Kihwan Kim, Jinwei Gu, Yasutaka Furukawa, and Jan Kautz.
\newblock Planercnn: 3d plane detection and reconstruction from a single image.
\newblock In {\em Proceedings of the IEEE/CVF Conference on Computer Vision and
  Pattern Recognition}, pages 4450--4459, 2019.

\bibitem{liu2018planenet}
Chen Liu, Jimei Yang, Duygu Ceylan, Ersin Yumer, and Yasutaka Furukawa.
\newblock Planenet: Piece-wise planar reconstruction from a single rgb image.
\newblock In {\em Proceedings of the IEEE Conference on Computer Vision and
  Pattern Recognition}, pages 2579--2588, 2018.

\bibitem{liu2022planemvs}
Jiachen Liu, Pan Ji, Nitin Bansal, Changjiang Cai, Qingan Yan, Xiaolei Huang,
  and Yi Xu.
\newblock Planemvs: 3d plane reconstruction from multi-view stereo.
\newblock In {\em Proceedings of the IEEE/CVF Conference on Computer Vision and
  Pattern Recognition}, pages 8665--8675, 2022.

\bibitem{luvizon2021adaptive}
Diogo~C Luvizon, Gustavo Sutter~P Carvalho, Andreza~A dos Santos, Jhonatas~S
  Conceicao, Jose~L Flores-Campana, Luis~GL Decker, Marcos~R Souza, Helio
  Pedrini, Antonio Joia, and Otavio~AB Penatti.
\newblock Adaptive multiplane image generation from a single internet picture.
\newblock In {\em Proceedings of the IEEE/CVF Winter Conference on Applications
  of Computer Vision}, pages 2556--2565, 2021.

\bibitem{martin2021nerf}
Ricardo Martin-Brualla, Noha Radwan, Mehdi~SM Sajjadi, Jonathan~T Barron,
  Alexey Dosovitskiy, and Daniel Duckworth.
\newblock Nerf in the wild: Neural radiance fields for unconstrained photo
  collections.
\newblock In {\em Proceedings of the IEEE/CVF Conference on Computer Vision and
  Pattern Recognition}, pages 7210--7219, 2021.

\bibitem{mildenhall2019llff}
Ben Mildenhall, Pratul~P. Srinivasan, Rodrigo Ortiz-Cayon, Nima~Khademi
  Kalantari, Ravi Ramamoorthi, Ren Ng, and Abhishek Kar.
\newblock Local light field fusion: Practical view synthesis with prescriptive
  sampling guidelines.
\newblock {\em ACM Transactions on Graphics (TOG)}, 2019.

\bibitem{mildenhall2020nerf}
Ben Mildenhall, Pratul~P. Srinivasan, Matthew Tancik, Jonathan~T. Barron, Ravi
  Ramamoorthi, and Ren Ng.
\newblock Nerf: Representing scenes as neural radiance fields for view
  synthesis.
\newblock In {\em ECCV}, 2020.

\bibitem{milletari2016v}
Fausto Milletari, Nassir Navab, and Seyed-Ahmad Ahmadi.
\newblock V-net: Fully convolutional neural networks for volumetric medical
  image segmentation.
\newblock In {\em 2016 fourth international conference on 3D vision (3DV)},
  pages 565--571. IEEE, 2016.

\bibitem{muller2022instant}
Thomas M{\"u}ller, Alex Evans, Christoph Schied, and Alexander Keller.
\newblock Instant neural graphics primitives with a multiresolution hash
  encoding.
\newblock {\em ACM Transactions on Graphics (ToG)}, 41(4):1--15, 2022.

\bibitem{roessle2022dense}
Barbara Roessle, Jonathan~T Barron, Ben Mildenhall, Pratul~P Srinivasan, and
  Matthias Nie{\ss}ner.
\newblock Dense depth priors for neural radiance fields from sparse input
  views.
\newblock In {\em Proceedings of the IEEE/CVF Conference on Computer Vision and
  Pattern Recognition}, pages 12892--12901, 2022.

\bibitem{shade1998layered}
Jonathan Shade, Steven Gortler, Li-wei He, and Richard Szeliski.
\newblock Layered depth images.
\newblock In {\em Proceedings of the 25th annual conference on Computer
  graphics and interactive techniques}, pages 231--242, 1998.

\bibitem{shih20203d}
Meng-Li Shih, Shih-Yang Su, Johannes Kopf, and Jia-Bin Huang.
\newblock 3d photography using context-aware layered depth inpainting.
\newblock In {\em Proceedings of the IEEE/CVF Conference on Computer Vision and
  Pattern Recognition}, pages 8028--8038, 2020.

\bibitem{silberman2012indoor}
Nathan Silberman, Derek Hoiem, Pushmeet Kohli, and Rob Fergus.
\newblock Indoor segmentation and support inference from rgbd images.
\newblock In {\em European conference on computer vision}, pages 746--760.
  Springer, 2012.

\bibitem{sinha2009piecewise}
Sudipta Sinha, Drew Steedly, and Rick Szeliski.
\newblock Piecewise planar stereo for image-based rendering.
\newblock In {\em 2009 International Conference on Computer Vision}, pages
  1881--1888, 2009.

\bibitem{srinivasan2019pushing}
Pratul~P Srinivasan, Richard Tucker, Jonathan~T Barron, Ravi Ramamoorthi, Ren
  Ng, and Noah Snavely.
\newblock Pushing the boundaries of view extrapolation with multiplane images.
\newblock In {\em Proceedings of the IEEE/CVF Conference on Computer Vision and
  Pattern Recognition}, pages 175--184, 2019.

\bibitem{sun2022direct}
Cheng Sun, Min Sun, and Hwann-Tzong Chen.
\newblock Direct voxel grid optimization: Super-fast convergence for radiance
  fields reconstruction.
\newblock In {\em Proceedings of the IEEE/CVF Conference on Computer Vision and
  Pattern Recognition}, pages 5459--5469, 2022.

\bibitem{tan2021planetr}
Bin Tan, Nan Xue, Song Bai, Tianfu Wu, and Gui-Song Xia.
\newblock Planetr: Structure-guided transformers for 3d plane recovery.
\newblock In {\em Proceedings of the IEEE/CVF International Conference on
  Computer Vision}, pages 4186--4195, 2021.

\bibitem{trevithick2021grf}
Alex Trevithick and Bo Yang.
\newblock Grf: Learning a general radiance field for 3d representation and
  rendering.
\newblock In {\em Proceedings of the IEEE/CVF International Conference on
  Computer Vision}, pages 15182--15192, 2021.

\bibitem{tucker2020single}
Richard Tucker and Noah Snavely.
\newblock Single-view view synthesis with multiplane images.
\newblock In {\em Proceedings of the IEEE/CVF Conference on Computer Vision and
  Pattern Recognition}, pages 551--560, 2020.

\bibitem{tulsiani2018layer}
Shubham Tulsiani, Richard Tucker, and Noah Snavely.
\newblock Layer-structured 3d scene inference via view synthesis.
\newblock In {\em Proceedings of the European Conference on Computer Vision
  (ECCV)}, pages 302--317, 2018.

\bibitem{wei2021nerfingmvs}
Yi Wei, Shaohui Liu, Yongming Rao, Wang Zhao, Jiwen Lu, and Jie Zhou.
\newblock Nerfingmvs: Guided optimization of neural radiance fields for indoor
  multi-view stereo.
\newblock In {\em Proceedings of the IEEE/CVF International Conference on
  Computer Vision}, pages 5610--5619, 2021.

\bibitem{wiles2020synsin}
Olivia Wiles, Georgia Gkioxari, Richard Szeliski, and Justin Johnson.
\newblock Synsin: End-to-end view synthesis from a single image.
\newblock In {\em Proceedings of the IEEE/CVF Conference on Computer Vision and
  Pattern Recognition}, pages 7467--7477, 2020.

\bibitem{xie2022planarrecon}
Yiming Xie, Matheus Gadelha, Fengting Yang, Xiaowei Zhou, and Huaizu Jiang.
\newblock Planarrecon: Real-time 3d plane detection and reconstruction from
  posed monocular videos.
\newblock In {\em Proceedings of the IEEE/CVF Conference on Computer Vision and
  Pattern Recognition}, pages 6219--6228, 2022.

\bibitem{xu2022sinnerf}
Dejia Xu, Yifan Jiang, Peihao Wang, Zhiwen Fan, Humphrey Shi, and Zhangyang
  Wang.
\newblock Sinnerf: Training neural radiance fields on complex scenes from a
  single image.
\newblock {\em arXiv preprint arXiv:2204.00928}, 2022.

\bibitem{yu2021pixelnerf}
Alex Yu, Vickie Ye, Matthew Tancik, and Angjoo Kanazawa.
\newblock pixelnerf: Neural radiance fields from one or few images.
\newblock In {\em Proceedings of the IEEE/CVF Conference on Computer Vision and
  Pattern Recognition}, pages 4578--4587, 2021.

\bibitem{yu2019single}
Zehao Yu, Jia Zheng, Dongze Lian, Zihan Zhou, and Shenghua Gao.
\newblock Single-image piece-wise planar 3d reconstruction via associative
  embedding.
\newblock In {\em Proceedings of the IEEE/CVF Conference on Computer Vision and
  Pattern Recognition}, pages 1029--1037, 2019.

\bibitem{zhou2018stereo}
Tinghui Zhou, Richard Tucker, John Flynn, Graham Fyffe, and Noah Snavely.
\newblock Stereo magnification: learning view synthesis using multiplane
  images.
\newblock {\em ACM Transactions on Graphics (TOG)}, 37(4):1--12, 2018.

\bibitem{zhou2016view}
Tinghui Zhou, Shubham Tulsiani, Weilun Sun, Jitendra Malik, and Alexei~A Efros.
\newblock View synthesis by appearance flow.
\newblock In {\em European conference on computer vision}, pages 286--301.
  Springer, 2016.

\end{thebibliography}
